\documentclass[conference]{IEEEtran}
\usepackage{cite}
\usepackage{amsmath,amssymb,amsfonts}
\usepackage{algorithmicx}
\usepackage{graphicx}
\usepackage{textcomp}
\usepackage{xcolor}
\def\BibTeX{{\rm B\kern-.05em{\sc i\kern-.025em b}\kern-.08em
    T\kern-.1667em\lower.7ex\hbox{E}\kern-.125emX}}
    
\usepackage{subcaption}
\usepackage{comment}
\usepackage{hyperref} 
\hypersetup{
    colorlinks=true,
    linkcolor=blue,
    filecolor=blue,      
    urlcolor=blue,
    citecolor=blue
}

\usepackage{todonotes}
\usepackage{multirow}
\graphicspath{{figures/}}
    \DeclareGraphicsExtensions{.pdf,.eps,.png,.jpg}
\usepackage{bm}	
\usepackage{booktabs}
\usepackage[noabbrev,capitalise]{cleveref} 
\crefname{equation}{}{}	
\usepackage{tikz}
\usepackage{upgreek, tikz, standalone}
\usepackage[utf8]{inputenc}
\usepackage[acronym]{glossaries}

\makeatletter
\newcommand{\linebreakand}{%
  \end{@IEEEauthorhalign}
  \hfill\mbox{}\par
  \mbox{}\hfill\begin{@IEEEauthorhalign}
}
\makeatother

\usepackage[nolist]{acronym}
\usepackage[english]{babel}

\begin{acronym}[DRAM]
    \acro{AI}{artificial intelligence}
    \acrodefindefinite{AI}{an}{an}
    \acro{ASIC}{application-specific integrated circuit}
    \acrodefindefinite{ASIC}{an}{an}
    \acro{BD}{bounded delay}
    \acro{BNN}{binarized neural network}
    \acro{CNN}{convolutional neural network}
    \acro{CTM}{convolutional Tsetlin machine}
    \acro{FSM}{finite state machine}
    \acro{HV}{hypervector}
    \acro{HVTM}{hypervector Tsetlin machine}
    \acro{LA}{learning automaton}
    \acrodefplural{LA}{learning automata}
    \acrodefindefinite{LA}{an}{a}
    \acro{ML}{machine learning}
    \acrodefindefinite{ML}{an}{a}
    \acro{NLP}{natural language processing}
    \acro{TA}{Tsetlin automaton}
    \acrodefplural{TA}{Tsetlin automata}
    \acro{TAT}{Tsetlin automaton team}
    \acro{TM}{Tsetlin machine}
    \acro{RTM}{regression Tsetlin machine}
    \acro{RbE}{Reasoning by Elimination}
    \acro{HVSize}{Hypervector size}
    \acro{NBits}{number of projection Booleans}
    \acro{HVTA}{Hypervector Tsetlin Automata}
\end{acronym}

\usepackage{algpseudocode}

\algnewcommand{\LineComment}[1]{\State \(\triangleright\) #1}    
    
\begin{document}

\title{Exploring Effects of Hyperdimensional Vectors for Tsetlin Machines}

\author{
\IEEEauthorblockN{Vojtech Halenka}
\IEEEauthorblockA{\textit{Department of ICT} \\
\textit{University of Agder}\\
Grimstad, Norway \\
vojtech.halenka@uia.no}
\and
\IEEEauthorblockN{Ahmed K. Kadhim}
\IEEEauthorblockA{\textit{Department of ICT} \\
\textit{University of Agder}\\
Grimstad, Norway \\
ahmed.k.kadhim@uia.no}
\and
\IEEEauthorblockN{Paul F. A. Clarke}
\IEEEauthorblockA{\textit{Department of ICT} \\
\textit{University of Agder}\\
Grimstad, Norway \\
paul.clarke@uia.no}
\and
\IEEEauthorblockN{Bimal Bhattarai}
\IEEEauthorblockA{\textit{Department of ICT} \\
\textit{University of Agder}\\
Grimstad, Norway \\
bimal.bhattarai@uia.no}
\and
\IEEEauthorblockN{Rupsa Saha}
\IEEEauthorblockA{\textit{Department of ICT} \\
\textit{University of Agder}\\
Grimstad, Norway \\
rupsa.saha@uia.no}
\linebreakand
\IEEEauthorblockN{Ole-Christoffer Granmo}
\IEEEauthorblockA{\textit{Department of ICT} \\
\textit{University of Agder}\\
Grimstad, Norway \\
ole.granmo@uia.no}
\and
\IEEEauthorblockN{Lei Jiao}
\IEEEauthorblockA{\textit{Department of ICT} \\
\textit{University of Agder}\\
Grimstad, Norway \\
lei.jiao@uia.no}
\and
\IEEEauthorblockN{Per-Arne Andersen}
\IEEEauthorblockA{\textit{Department of ICT} \\
\textit{University of Agder}\\
Grimstad, Norway \\
per.andersen@uia.no}
}
\maketitle

\begin{abstract}
Tsetlin machines (TMs) have been successful in several application domains, operating with high efficiency on Boolean representations of the input data.
However, Booleanizing complex data structures such as sequences, graphs, images, signal spectra, chemical compounds, and natural language is not trivial. In this paper, we propose a \ac{HV} based method for expressing arbitrarily large sets of concepts associated with any input data. Using a hyperdimensional space to build vectors drastically expands the capacity and flexibility of the TM. We demonstrate how images, chemical compounds, and natural language text are encoded according to the proposed method, and how the resulting \ac{HV}-powered TM can achieve significantly higher accuracy and faster learning on well-known benchmarks. 
Our results open up a new research direction for TMs, namely how to expand and exploit the benefits of operating in hyperspace, including new booleanization strategies, optimization of TM inference and learning, as well as new TM applications.
\end{abstract}
\section{Introduction}\label{sec:intro}


The success of an AI model depends on the internal representation of the data that the model considers. An appropriate representation allows the model to exploit the correct juxtaposition of the features available in the data to attain the best possible performance it is capable of. Moreover, in the case of interpretable AI models, this representation should also offer a one-to-one mapping between human-understandable features and features that are advantageous to the model. For \acp{TM}, designing such feature spaces is referred to as data Boolanization, which is notoriously challenging for complex high-dimensional data, like sequences, graphs, images,
signal spectra, and natural language. The reason is that the features must be natural building blocks for creating AND-rules that are both interpretable and accurate.

We here propose the \ac{HVTM}, which extends the capabilities of \acp{TM} by using hyperdimensional computing~\cite{schlegel2022comparison} with \ac{HV} tokens.  So-called binding operations combine tokens into more complex structures, while bundling operations assemble the resulting structures into a rich representation of the data. Accordingly, we preserve the uniqueness of the data while improving decision-making accuracy and efficiency.

The paper contributions can be summarized as follows:
\begin{itemize}
    \item Our work introduces novel Booleanization strategies and hyperdimensional data analysis techniques based on the \ac{TM}.
    \item We show how the \ac{TM} can operate effectively in sparse hyperspaces, leveraging the capability of \acp{TM} to find non-linear patterns in high-dimensional spaces.
    \item We expose how a unique hyperparameter configuration where specificty $s$ is set to $1$, henceforth termed \ac{RbE} ~\cite{yadav2022robust}, is particularly suited for sparse hyperspaces.
    \item We further demonstrate the improvements achieved by the \ac{HVTM} via experiments on multiple datasets in the domains of image processing, chemical structures identification and \ac{NLP}.
    \item In each experiment, the \ac{HVTM} is compared to the standard \ac{TM}, to demonstrate their differences under the effect of various hyperparameter settings.
\end{itemize}

The paper is organized as follows. In Section~II, we provide the details of \ac{HVTM}, which we evaluate rigorously in Section~III, contrasting it against the vanilla \ac{TM}. Section~IV concludes the paper by summarizing our findings and providing pointers to further work.

\section{Hypervector Tsetlin Machine}\label{sec:tm_hv}
\begin{figure}[h!]
\centering
\includegraphics[width=1\linewidth]{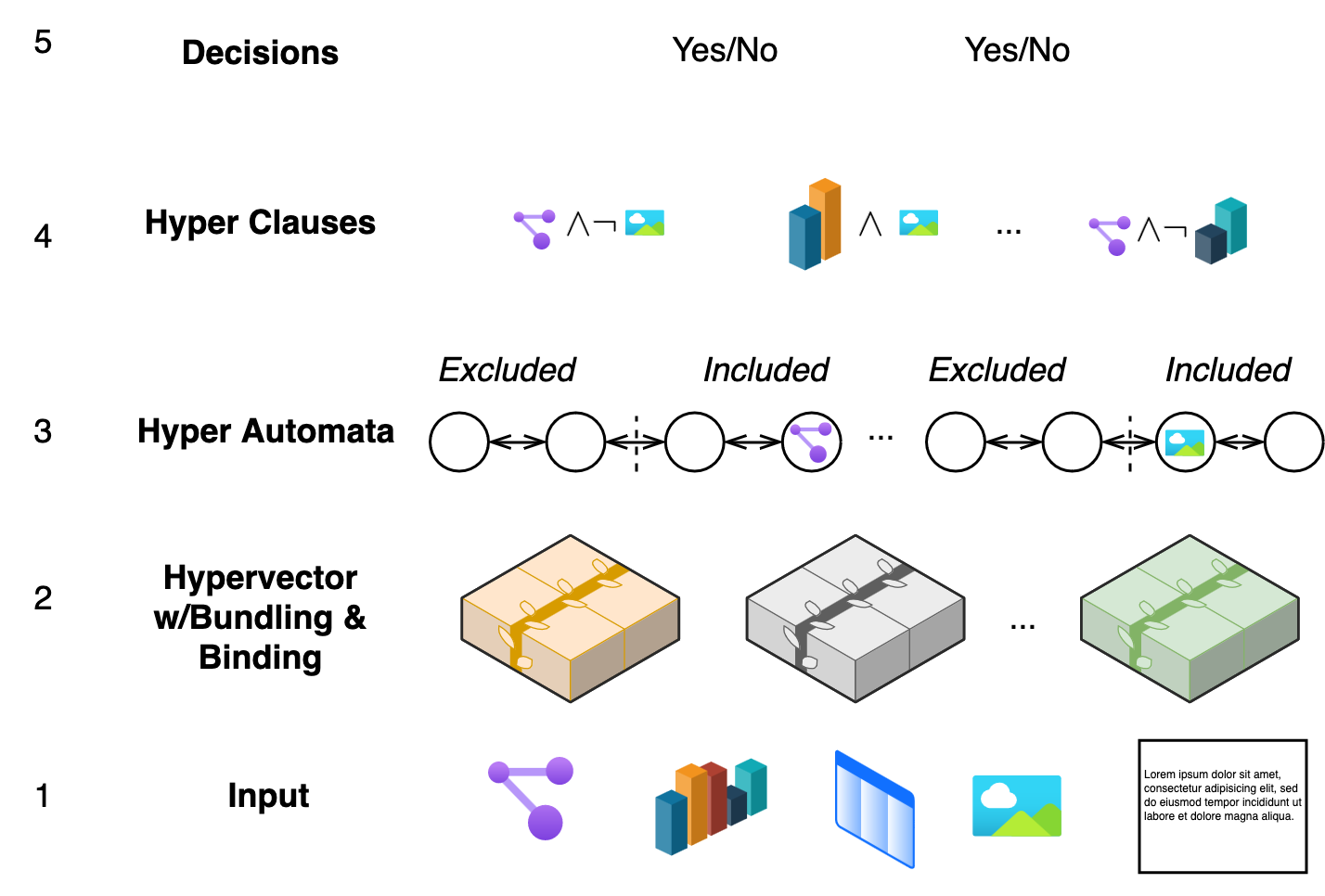}
\caption{Tsetlin machine operating in hyperspace}\label{figure:hypervector_tm}
\label{hypervector-tm-workflow}
\end{figure}

A \ac{HVTM} extends the foundational principles of a standard \ac{TM} into high-dimensional hyperspaces. Unlike traditional \acp{TM} that process Boolean inputs, a \ac{HVTM} operates on Hyperboolean \cite{HyperbooleanAlgebra} inputs -- complex constructs that emerge from operations within a hyperspace, effectively bypassing the limitations of a fixed-sized feature vector. The workflow of \ac{HVTM}, as visualized in Figure \ref{figure:hypervector_tm}, consists of the following steps, elaborated below.

\subsection{Workflow of Hypervector Tsetlin Machine}

\begin{enumerate}
    \item \textbf{Input Tokenization}:  Input data of various forms, such as text, images, or any complex data structure, is assigned a unique \ac{HV} token, which is generated randomly.

    \item \textbf{Bundling and Binding}
    Using bundling and binding techniques, we combine the tokens into high-dimensional vectors. \acp{HV} are thus made of hyper literals, each of which represents complex features of the input data.
    
    \item \textbf{Hyper Automata}: Encoded \acp{HV} are then processed by a layer of hyper automata. Each hyper automaton functions identically to a \ac{TA}, making decisions on whether to include or exclude a particular hyper literal. 
    
    \item \textbf{Hyper Clauses}: Hyper clauses are formed through the process of collective decision-making by the hyper automata. Each hyper clause represents a pattern or a rule that contributes to the decision-making process, and are the high-dimensional equivalent of conjunctive clauses in a standard \ac{TM}.
    
    \item \textbf{Decisions}: The hyper clauses  finally vote for or against Yes/No output decisions according to how they match the input sample.
    
\end{enumerate}

Following the above procedure, the \ac{HVTM} is capable of handling high-dimensional data, making it applicable to a wider range of complex pattern recognition tasks, compared to the standard \ac{TM}. The hyperspace operation allows for capturing intricate data relationships, potentially leading to more nuanced and sophisticated decision-making. How the \ac{HVTM} processes \acp{HV}, in contrast to a regular \ac{TM} operating on vectors, is explained in detail in the following section.

\subsection{Hypervectors}

A hyperspace gets arbitrarily high dimensionality by representing each dimension with a fixed-sized \ac{HV}, typically of high dimensionality. Because of the fixed  \ac{HV} size, the hyperspace representation is lossy. However, increasing the \ac{HV} size reduces the lossiness to any degree. Accordingly, \acp{HV}  allows for a more complex and nuanced representation of data~\cite{aygun2023learning}.

In our approach, we employ sparse \acp{HV}, which are binary/Boolean~\cite{schlegel2022comparison}. A sparse \ac{HV} consists of a large fixed number of Booleans, e.g., $1024$, referred to as \ac{HVSize} in the following. The \ac{HV} gets a unique signature by setting a small number of the Booleans to \emph{True}, randomly selected. These are referred to as the \ac{NBits}. We call the resulting \ac{HV} as a token, representing a dimension in hyperspace.  So, our first step is tokenization of the data, where each unique token forms a dimension. This process is pivotal in imparting complex meanings to the Booleans processed by the \ac{HVTM}.

The synthesis of complex structures with sparse \acp{HV} is facilitated by operations known as Binding and Bundling. Binding involves a cyclic shift, effectively creating a new dimension in hyperspace, while Bundling is achieved through the logical OR operation, populating the hyperspace. This method of synthesis is crucial for creating new, unified hypervectors from existing ones \cite{yu2023understanding}.

\subsubsection{Creation of a hypervector}

\begin{figure}[ht]
\centering
\includegraphics[width=1\linewidth]{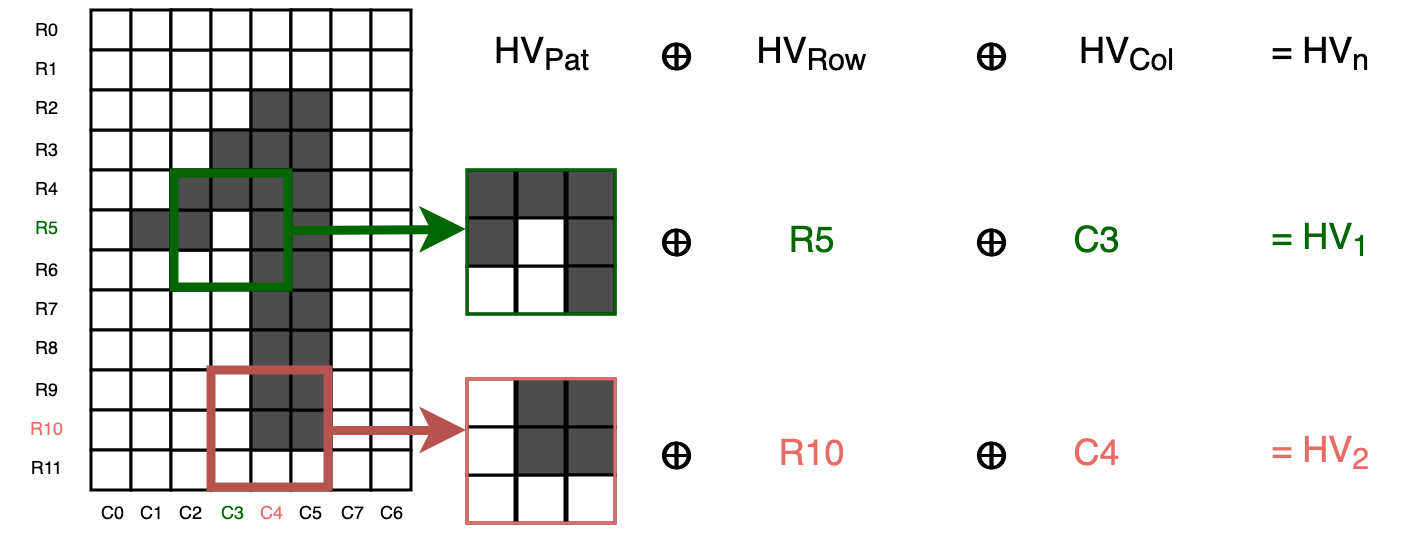}
\caption{Creation of a Hypervector from 3 different hypervector tokens, bundling Patch with its respective Row and Column}\label{figure:img_HV_creation}
\end{figure}

Figure \ref{figure:img_HV_creation} illustrates the process of creating a \ac{HV} within the \ac{HVTM} framework, using an image example. The process begins by extracting a `Patch' \ac{HV} (HVPat) from the input image, which is represented as a grid of binary values, with black squares indicating a value of 1 and white squares a value of 0. This Patch is then bound with its corresponding `Row' (HVRow) and `Column' (HVCol) \acp{HV}, which are extracted based on the position of the Patch within the image (Row 5, Column 3 for HV1, and Row 10, Column 4 for HV2 in this example). The bundling operation involves a cyclic shift and logical OR, which combines these individual \acp{HV} to form a new, unified \ac{HV} representing both the pattern and its position within the input space. For instance, HV1 is formed by bundling HVPat with R5 and C3, and similarly, HV2 with R10 and C4. This step-by-step combination of \acp{HV} is pivotal in encoding the spatial information of the Patch within the larger context of the image, thereby enhancing the data representation for subsequent processing by the \ac{HVTM}.

\subsubsection{Explainability} 
\begin{figure}[ht]
\centering
\includegraphics[width=1\linewidth]{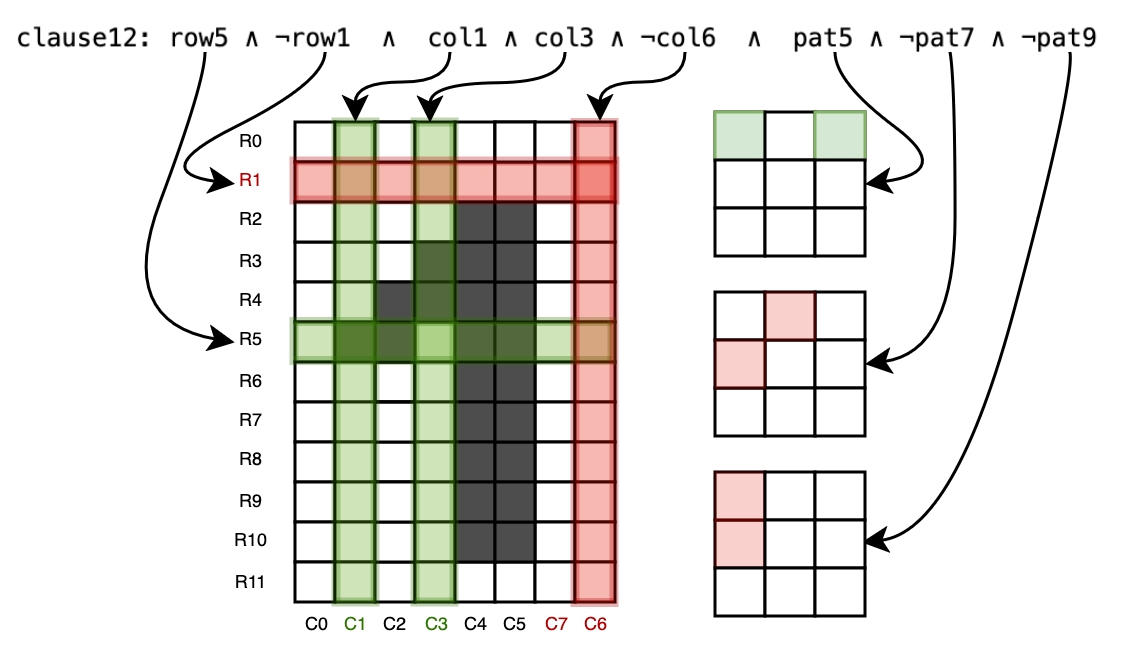}
\caption{Interpretation of exported clauses after training, with hyperliterals representing rows, columns and different patches}\label{fig:IMG_explainability}
\end{figure}

The \ac{HVTM}, rooted in the principles of the \ac{TM}, maintains its built-in capacity for understandable decision-making. Model's reasoning can be decoded by breaking down hyperclauses, tracking individual hyperliterals carrying their specific meaning assigned during the encoding process. For example, in the Figure \ref{fig:IMG_explainability} we follow a clause extracted from the \ac{HVTM} trained on the MNIST dataset. This clause contains both votes pro and against certain rows, columns and patches, which are the hyperfeatures that \ac{HVTM} considers. 

In this context, the \ac{HVTM} remains transparent and interpretable, allowing users to understand model decisions.

\subsubsection{Dimensionality and Storage Limitations}
Given the high-dimensional nature of \acp{HV}, the capacity of the \ac{HVTM} to encode and differentiate between tokens is inherently linked to the hypervector size, denoted as \(D\). This size restricts the maximum number of unique tokens that can be stored distinctly within the \ac{HVTM}'s framework, where each token requires a certain quota of 'space' or dimensions for its representation without ambiguity. The theoretical capacity \(C\) of a \ac{HV} to store unique tokens can be approximated, under ideal conditions, by the formula:
\[
C \approx \frac{D!}{S!(D-S)!},
\]
where \(S\) represents the sparsity or the number of non-zero elements.

\subsubsection{Projection Overlaps and Robustness}
The token projection process aims to minimize overlaps. Nonetheless, when the diversity of input data or the token count surpasses the hypervector's capacity, overlaps become inevitable. This reduces the HVTM's robustness against overlaps, inversely proportional to the number of projection bits \(P\). The likelihood \(L\) of overlaps increases with the number of tokens \(T\) relative to the hypervector size \(D\), as shown by:
\[
L = 1 - \left(1 - \frac{1}{D^P}\right)^T.
\]

\subsubsection{Scalability and Computational Complexity}
The scalability of HVTM is constrained by the dimensional size of hypervectors and the complexity of the Binding and Bundling operations. High-dimensional spaces necessitate more computational resources, potentially limiting the practical deployment of HVTM for large datasets or in resource-constrained environments.

To mitigate these limitations, further research is needed to optimize hypervector representations and develop more efficient algorithms for high-dimensional space management. This includes exploring adaptive or dynamic dimensionality adjustment mechanisms to balance storage capacity, token generation, and computational efficiency. Most promising research seems to be in Hardware Accelerators. 

Understanding and addressing these limitations is crucial for advancing HVTM research towards more robust, scalable, and efficient models that leverage the benefits of high-dimensional computing while mitigating inherent challenges.

\subsection{Internal mechanism of HVTM }
A \ac{HVTM} undertakes the task of binary classification for an input boolean \ac{HV} $H=[h_1,\ldots,h_o]$, where each $h_i$ represents a boolean hyperfeature.
These hyperfeatures, along with their negated counterparts, form the hyperliterals pool, facilitating the \ac{HVTM}'s processing.

For each target class, the \ac{HVTM} constructs patterns using $n$ conjunctive hyperclauses, with $n$ being a predefined parameter by the user. Positive polarity is allocated to half of the clauses with an odd index, while negative polarity is assigned to the other half with an even index. The positive clauses carry information that supports the class, whereas the negative polarity clauses carry information that opposes the class. Each hyperclause, denoted as $C_j(H)$, is made up of a set of hyperliterals $L_j$ that are subsets of $L$, i.e.  $L_j \subset L$. The clause $C_j(H) = h_1 \land h_2 = h_1 h_2$, for example, comprises of the literals $L_j = \{h_1, h_2\}$, $\bar{L_j} = \{\neg h_1, \neg h_2\}$ and outputs $1$ if $h_1 = h_2 = 1$.

The clause output is a measure of the certainty that a certain clause may contribute to the right classification decision. The unit step function is used to integrate the clause outputs into a classification decision through summation and thresholding using the unit step function $u(v) = 1 ~\mathbf{if}~ v \ge 0 ~\mathbf{else}~ 0$:
\begin{equation}
\textstyle
\hat{y} = u\left(\sum_{j=1}^{n/2} C_j^+(H) - \sum_{j=1}^{n/2} C_j^-(H)\right).
\end{equation}
Namely, classification is done based on a majority vote, with the positive clauses voting for $y=1$ and the negative for $y=0$. 

\begin{figure}[ht]
\centering
\includegraphics[width=1\linewidth]{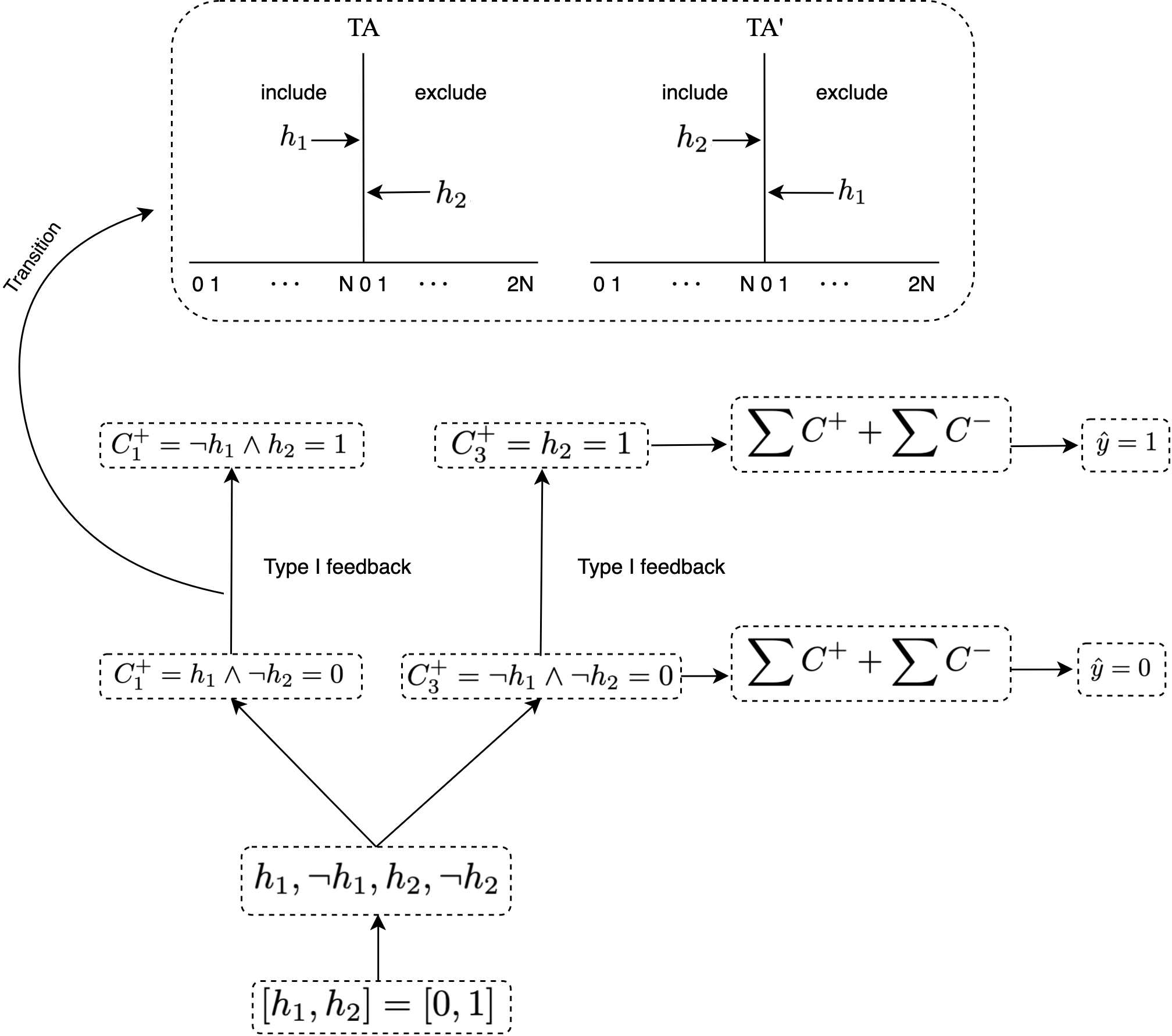}
\caption{The learning of Tsetlin Machine for a sample of XOR gate.}\label{figure:tm_architecture_basic}
\end{figure}


A hyperclause $C_j(H)$ is created by a group of \acp{HVTA} \cite{Tsetlin1961}, with each \ac{HVTA} selecting whether to \emph{Include} or \emph{Exclude} a specific hyperliteral $l_k$ in the clause. The feedback \ac{TA} receives in the form of Reward, Inaction, and Penalty is used to make decisions. \ac{TM} provides two types of feedback: Type I Feedback and Type II Feedback, and their mechanism of application for the \ac{HVTM} is the same as in a standard \ac{TM}~\cite{granmo2018tsetlin}.
\section{Empirical Results}\label{sec:empirical}

\begin{table}[ht!]
\centering
\setlength\tabcolsep{0pt} 
\label{performance}
\begin{tabular*}{0.4\textwidth}{@{\extracolsep{\fill}} lcccc}
\toprule
 & \multicolumn{4}{c}{Dataset Performance (\%)} \\
 & \multicolumn{1}{c}{TREC} & \multicolumn{1}{c}{HIV} & \multicolumn{1}{c}{IMDB} & \multicolumn{1}{c}{MNIST} \\ 
\midrule
HVTM      &95.6       &86.52   &89.67       & 98.13 \\
TM        &91.6       &85.57       &86.61       & 97.2  \\
\bottomrule
\end{tabular*}
\caption{Comparison of performance of \ac{HVTM} vs \ac{TM}, with identical parameters for both}
\label{tab:standard_vs_HVTM_results}
\end{table}



\subsection{Categorization of Natural Language Texts}
\label{nlp_experiments}


In this section, we detail our experiments with Sparse \ac{HV} to explore the \ac{HVTM} in \ac{NLP} in the context of the IMDB dataset~\cite{DBLP:journals/corr/abs-1909-12434} ($25000$ movie and series reviews), and the TREC dataset~\cite{voorhees2000building} ($5500$ question-answer pairs) fact-based question classification).

\begin{figure}
    \centering
    \includegraphics[width=1\linewidth]{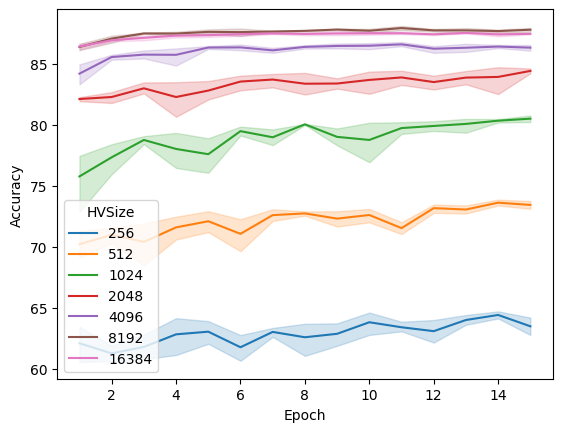}
    \caption{Sparse \ac{HV} with varying \ac{HVSize} for IMDB }
    \label{fig:imdb_epochvsacc_inc_hvsize_sparse}
\end{figure}

\begin{figure}
    \centering
    \includegraphics[width=1\linewidth]{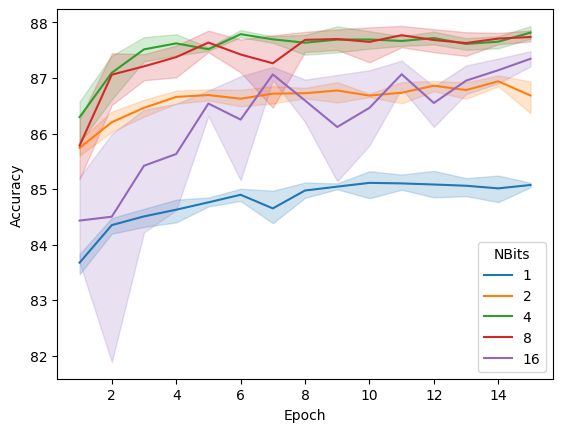} 
    \caption{Sparse \ac{HV} with varying \ac{NBits} for IMDB}
    \label{fig:imdb_epochvsacc_inc_projbits_sparse}
\end{figure}

\begin{figure}
    \centering
    \includegraphics[width=1\linewidth]{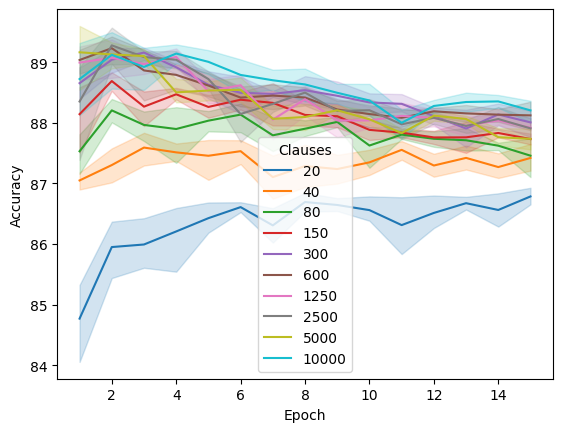}
    \caption{Sparse \ac{HV} with \ac{RbE} for IMDB}
    \label{fig:imdb_epochvsacc_RBE_sparse}
\end{figure}

\begin{figure}
    \centering
    \includegraphics[width=1\linewidth]{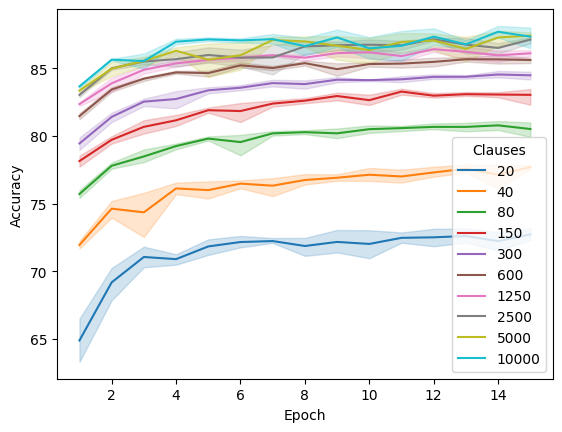}
    \caption{Sparse \ac{HV} without \ac{RbE} for IMDB}
    \label{fig:imdb_epochvsacc_woRBE_sparse}
\end{figure}

In IMDB, the highest accuracy rate achieved was 88.1\% with \ac{HVSize} of $8192$ and \ac{NBits} as $4$ (Fig.~\ref{fig:imdb_epochvsacc_inc_hvsize_sparse}). Fig.~\ref{fig:imdb_epochvsacc_inc_projbits_sparse} shows that an optimal number of projection bits positively affects the accuracy, reaching a maximum of $88$\% for $4$ bits with \ac{HVSize} of $10000$. Using \ac{RbE} further improves accuracy to $89.67$\% (Fig.~\ref{fig:imdb_epochvsacc_RBE_sparse}), while Fig.~\ref{fig:imdb_epochvsacc_woRBE_sparse} shows that without \ac{RbE} the maximum accuracy was 88.42\%.

\begin{figure}[h!]
\centering
\includegraphics[width=1\linewidth]{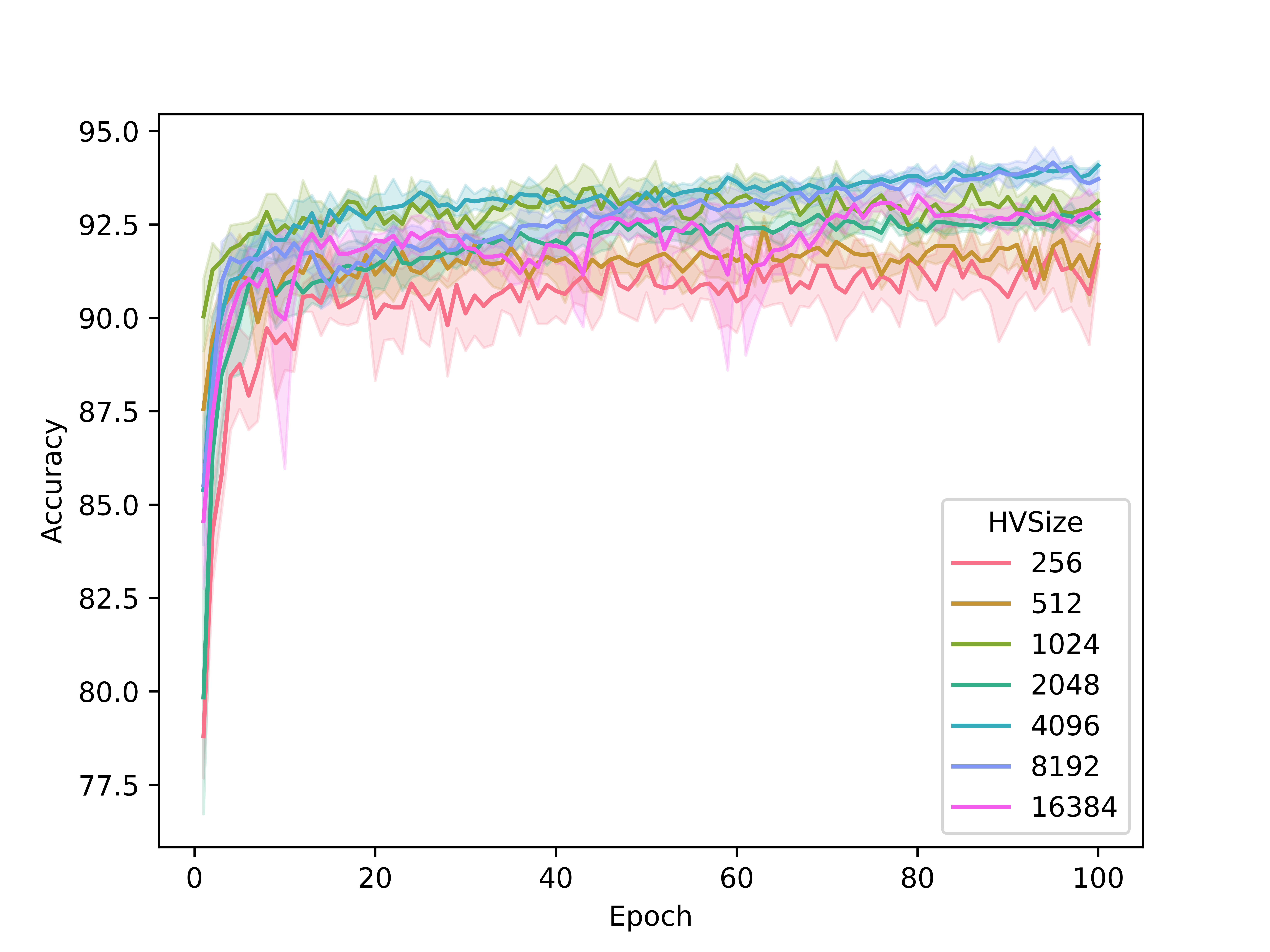}
\caption{Accuracy versus training epochs of sparse \ac{HVTM} for increasing \ac{HVSize} on TREC dataset. \ac{NBits} =  4.}\label{figure:nlp_trec_4bits}
\end{figure}

\begin{figure}[h!]
\centering
\includegraphics[width=1\linewidth]{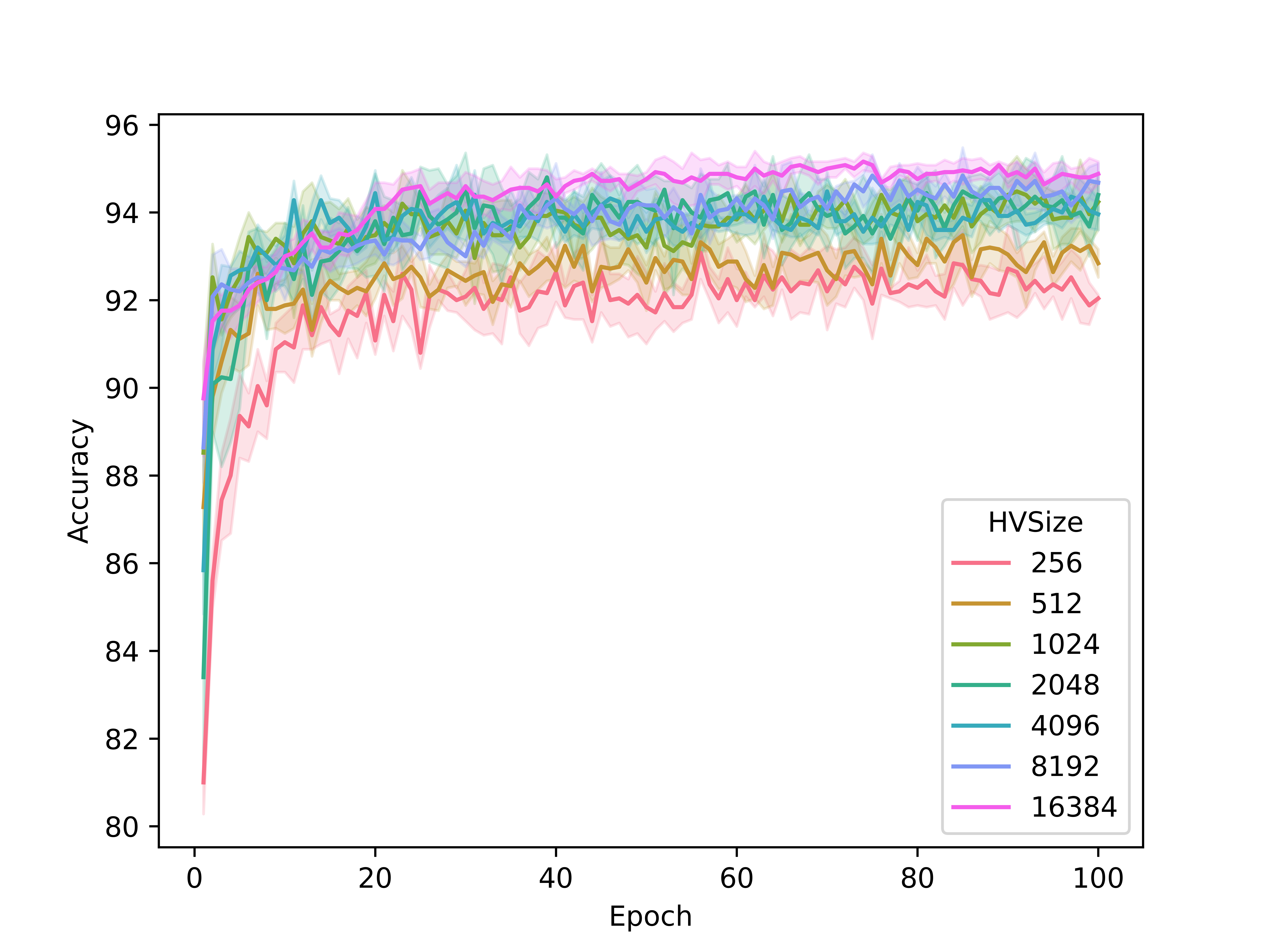}
\caption{Accuracy versus training epochs of sparse \ac{HVTM} for increasing \ac{HVSize} on TREC dataset. \ac{NBits} = 16}\label{figure:nlp_trec_16bits}
\end{figure}

\begin{figure}[h!]
\centering
\includegraphics[width=1\linewidth]{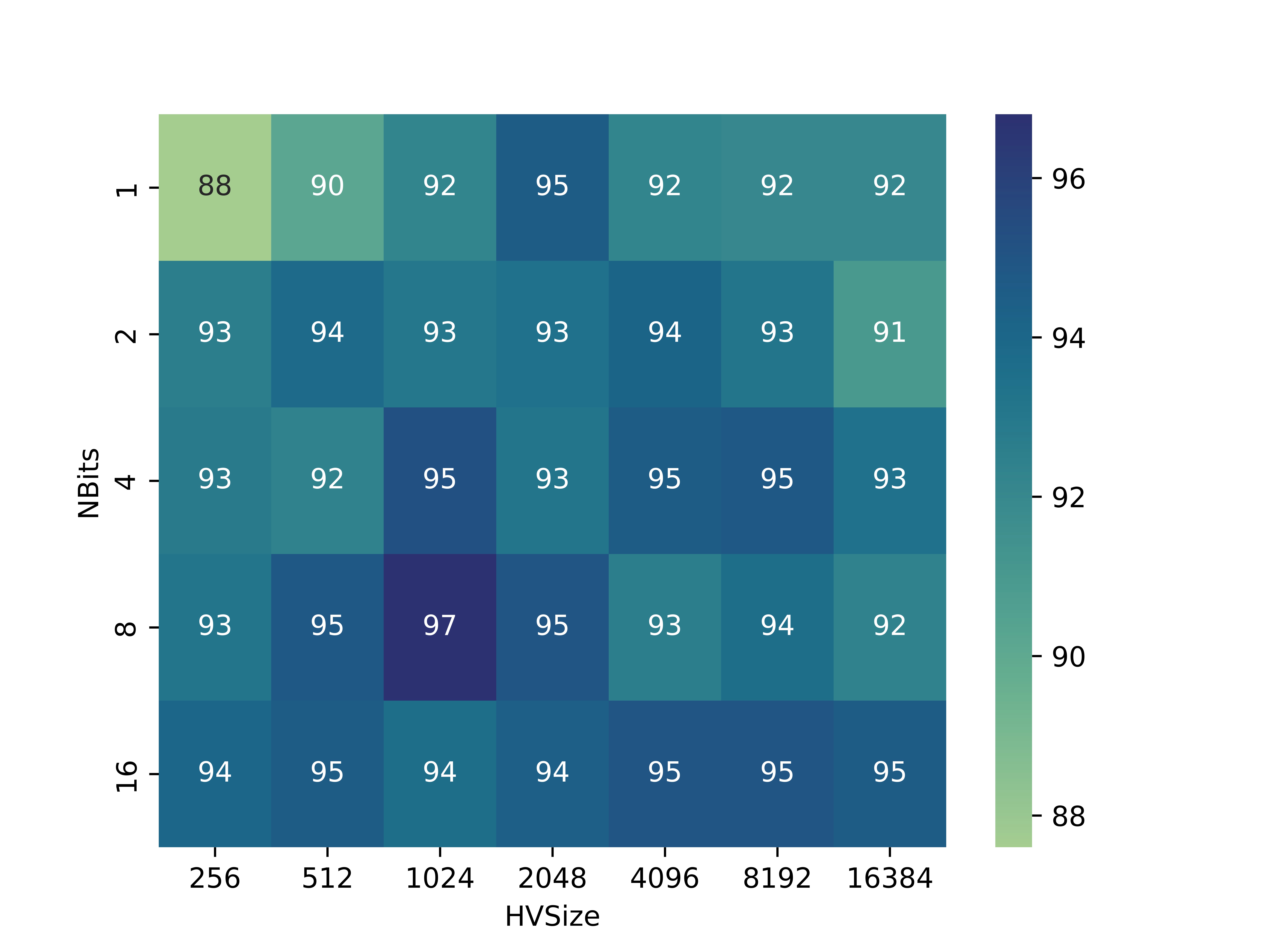}
\caption{Heatmap of accuracy for sparse HVTM as a function of HV Size and the number of activated bits on TREC Dataset}\label{figure:trec_sparse_heatmap}
\end{figure}

However, for TREC, highest accuracy of $96.8\%$ is achieved with a \ac{HVSize} of $1024$ and \ac{NBits} of $8$ given in Fig. \ref{figure:trec_sparse_heatmap}, which presents heatmap of maximum accuracy for each \ac{NBits} and \ac{HVSize}. As \ac{NBits} increases from $4$ (Fig. \ref{figure:nlp_trec_4bits}) to $16$ (Fig. \ref{figure:nlp_trec_16bits}), the \ac{HVTM} exhibits increased stability 
though the accuracy remains similar.\\

\begin{figure}[h!]
    \centering
    \includegraphics[width=1\linewidth]{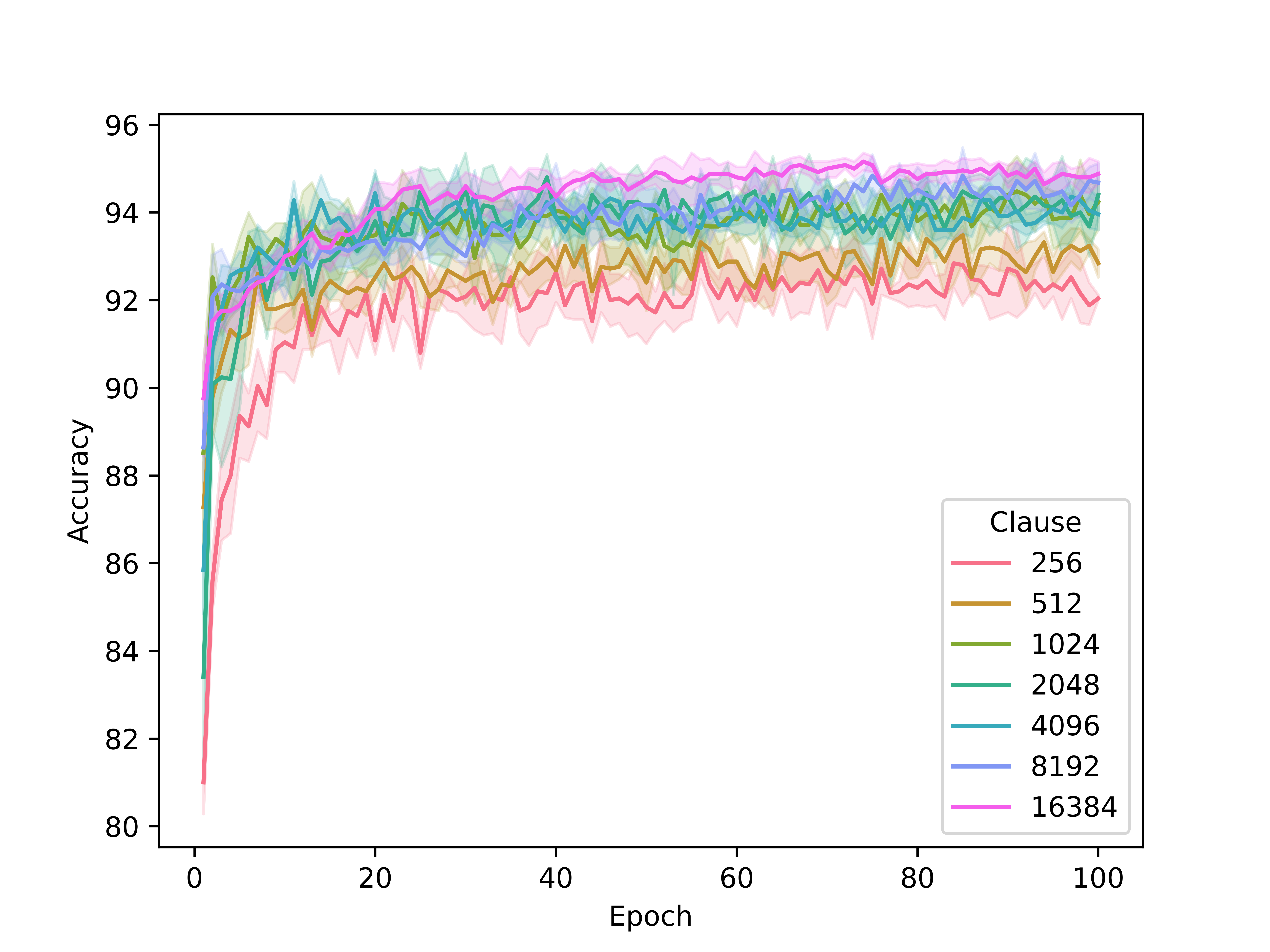}
    \caption{Graceful degradation by decrease in number of clauses on TREC dataset}
    \label{fig:trec_reasoning_clauses}
\end{figure}

\subsection{Classification of Compounds in Cheminformatics}

The HIV dataset, introduced by the Drug Therapeutics Program (DTP) AIDS Antiviral Screen, tests the ability to inhibit HIV replication of $40000$ compounds \cite{nih_hiv}. Compounds are evaluated as: confirmed inactive (CI), confirmed active (CA) and confirmed moderately active (CM). For this work, we simplify the problem into a binary classification by merging the latter two labels, resulting in classes Inactive (CI) and Active (CA + CM). The HIV dataset is imbalanced with large numbers of inactive compounds ($39$, $684$) and few actives ($1$, $443$).

The presence and absence of 2D sub-structures for compounds of the HIV dataset are encoded using Extended Connectivity Fingerprints (ECFP) with a vector length of $4096$ and circular sub-structure graph radius of 3 bond-lengths (ECFP6) \cite{rogers2010extended}. This initial feature set was used to build a conventional \ac{TM} model to which the \ac{HVTM} models are compared. 

\begin{figure}[h!]
\centering
\includegraphics[width=1\linewidth]{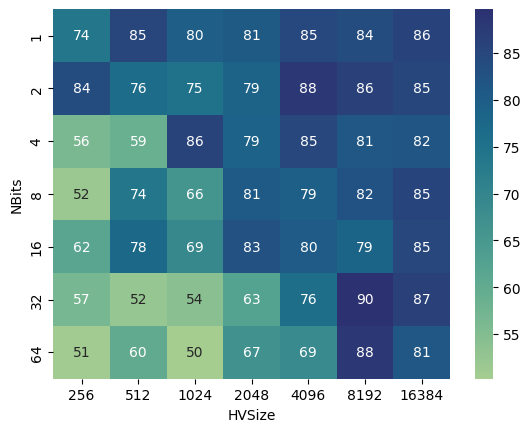}
\caption{Heatmap of BalAccuracy for \ac{HVTM} as a function of \acp{HVSize} and \ac{NBits} on HIV Dataset}\label{figure:chemHV_HVSize_NBits_BA}
\end{figure}

The ECFP6 sub-structure encodings were subsequently converted into \ac{HV} representations of differing \acp{HVSize} and \ac{NBits}. A given sub-structure bit position of the original ECFP6 features receives its own binary sparse \ac{HV}, the description of a compound is then completed by a bit-wise disjunction operation across all present bit positions \acp{HV}. For each \ac{HVSize} and \ac{NBits} combination, a \ac{HVTM} model was built and compared to each other via balanced accuracy (BalAccuracy) (Equation \ref{eq:balanced_accuracy}) \cite{5597285}. This process was repeated 5 times with the mean values reported in Fig. \ref{figure:chemHV_HVSize_NBits_BA}.

\begin{equation}
    Balanced Accuracy = \frac{\sum_{i}^{C} Recall_{i}}{N},
\label{eq:balanced_accuracy}
\end{equation}
where $i$ is a given class, $C$ is the set of all classes and $N$ is the total number of classes.

\begin{figure}[h!]
\centering
\includegraphics[width=1\linewidth]{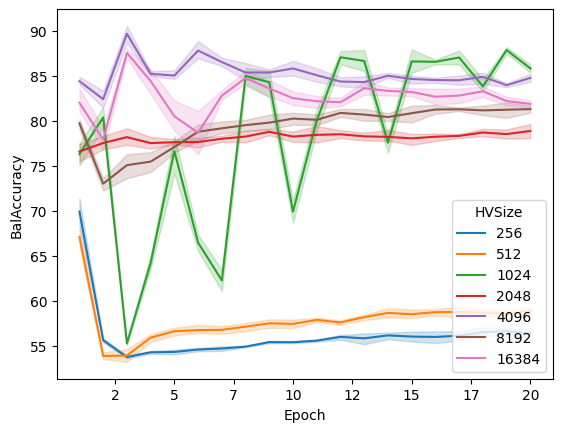}
\caption{Balanced accuracy over training epochs of \ac{HVTM}, for increasing \ac{HVSize}, on chemical structures of HIV dataset. \ac{NBits}=$4$.}\label{figure:chemHV_Acc_Epoch_NBits4}
\end{figure}

\begin{figure}[h!]
\centering
\includegraphics[width=1\linewidth]{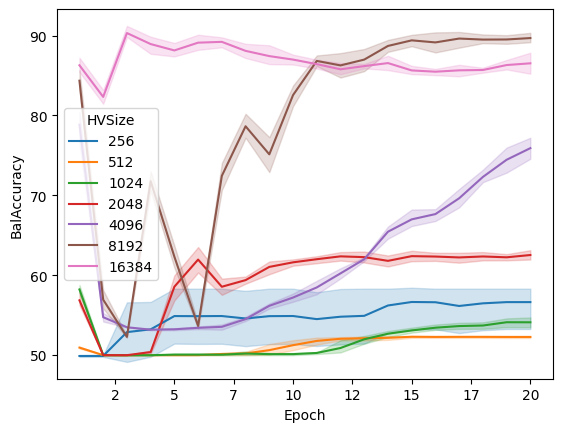}
\caption{Balanced accuracy versus training epochs of \ac{HVTM}, for increasing \acp{HVSize}, on HIV dataset. \ac{NBits}=$32$}\label{figure:chemHV_Acc_Epoch_NBits32}
\end{figure}

Fig. \ref{figure:chemHV_HVSize_NBits_BA} shows that usually, the average balanced accuracy for classifying compounds as HIV active, across all \ac{NBits}, increases with \ac{HVSize}. Deviations from this trend can be seen, especially for the combinations: $1024$ \ac{HVSize} with $4$ \ac{NBits}, $8192$ \ac{HVSize} and $32$ \ac{NBits}. We hypothesize that bit collisions in the \ac{HV} space allow the TM to create clauses which avail of sub-structure "groups" encoded in a single \ac{HV} Boolean position. These ideas are postulated by examining the balanced accuracy versus epoch curves (Fig. \ref{figure:chemHV_Acc_Epoch_NBits4} and \ref{figure:chemHV_Acc_Epoch_NBits32}) of the \ac{NBits} $4$ and $32$ respectively, and the corresponding cells in Fig. \ref{figure:chemHV_HVSize_NBits_BA}.

Fig. \ref{figure:chemHV_Acc_Epoch_NBits4} and Fig. \ref{figure:chemHV_Acc_Epoch_NBits32} exhibit similar patterns, where low \acp{HVSize} prevent \acp{HVTM} from mining frequent patterns, until a certain \acp{HVSize} is reached and learning is volatile but ultimately accurate. At higher \acp{HVSize}, patterns in \ac{HV} space were extracted quickly (after just 2 epochs) and with stable balanced accuracy ($86.52$\%), more than that of $85.57$\% achieved by the standard \ac{TM} (Table \ref{tab:standard_vs_HVTM_results}). In the case of \ac{NBits}=$32$, it takes a much larger \acp{HVSize} ($\geq 8192$) to cross this threshold into stable learning for \acp{HVTM} compared to \ac{NBits}=$4$ ($\geq 1024$). This is likely due to the greater number of bit collisions acting as noise when \ac{NBits} is higher. 

To reiterate, the high balanced accuracy achieved by the volatile HVTMs (NBits = $32$ and HVSize = $8192$, NBits = $4$ and HVSize = $1024$) after 20 epochs may be due to the grouping effect described earlier, where \acp{HVTM} use \ac{HV} bit collisions of multiple sub-structure positions to effectively group and discriminate between compound sub-structures with their limited clause pool. This "grouping" effect is just one of many potential hypotheses which will be examined in the future for chemical \ac{HVTM} models.
\subsection{Image Classification}
We conduct a series of experiments on the MNIST dataset~\cite{deng2012mnist} to investigate the impacts of \ac{HVSize}, \ac{NBits}, and clause count on classification accuracy. To mitigate stochastic variations, each experimental condition was evaluated using five or more ensemble runs.



\begin{figure}[ht]
\centering
\includegraphics[width=1\linewidth]{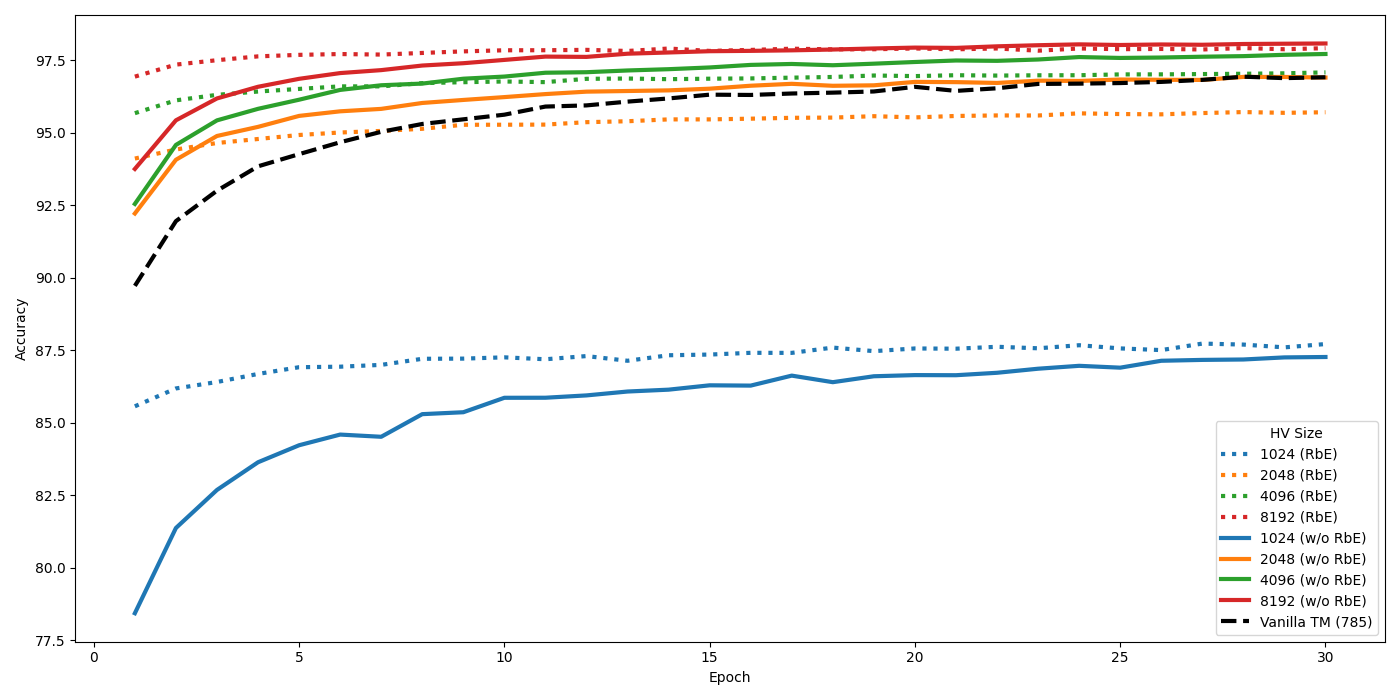}
\caption{Comparing \ac{HVTM} with \ac{RbE} with varying \ac{HVSize} and Vanilla TM 
}\label{fig:IMG_HV_size}
\end{figure}

However, enforcing overlaps between \ac{HV} tokens, for instance, aligning part of a token with specific columns or rows in the image (as in Figure \ref{figure:img_HV_creation}), enhances overall accuracy despite the same \ac{HVSize} (Fig. \ref{fig:IMG_HV_size} shows the baseline results by Vanilla TM are surpassed). 

Fig. \ref{fig:IMG_HV_size} also highlights the effect of \ac{HVSize}. When the size is smaller than the amount of information we are trying to compress (note the 1024 bit \ac{HVSize} in the figure), we get a lossy compression due to overlaps, resulting in a massive decrease in accuracy. 
Also observed in Figure \ref{fig:IMG_HV_size} is the effect of using \ac{RbE}. It seems to be faster in reaching higher accuracy in first few epochs, but eventually gets surpassed by the non-\ac{RbE} setup.

\begin{figure}[ht]
\centering
\includegraphics[width=1\linewidth]{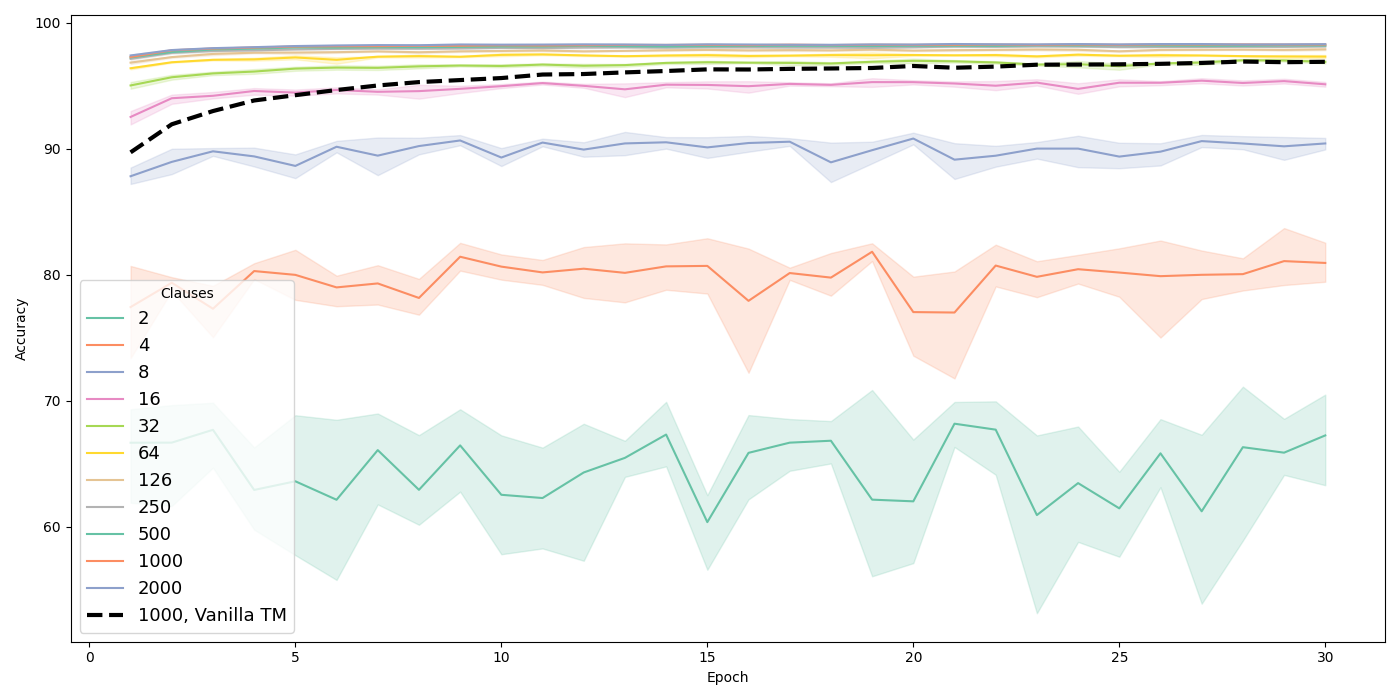}
\caption{Performance degradation due to reduced clause count in the \ac{HVTM} with \ac{RbE} on the MNIST dataset}\label{fig:IMG_Clauses}
\end{figure}

Lastly, we examine the effect of reducing the number of clauses in a \ac{HVTM}, which leads to graceful performance degradation (Fig. \ref{fig:IMG_Clauses}). Notably, even with a substantially more literals in the \ac{HVTM} ($16384$ literals) versus in the traditional TM ($785$ literals), the \ac{HVTM} achieves comparable performance with significantly fewer clauses.
In the next Fig.~\ref{fig:IMG_Clauses_det}, we see the detailed view of performance degradation due to reduced clause count in the HVTM on the MNIST dataset, employing \ac{RbE}. What's also interesting to note is the difference in variation among ensembles. The higher the number of clauses, the more stable is both the rise in performance and the performance itself.

\begin{figure}[ht]
\centering
\includegraphics[width=1\linewidth]{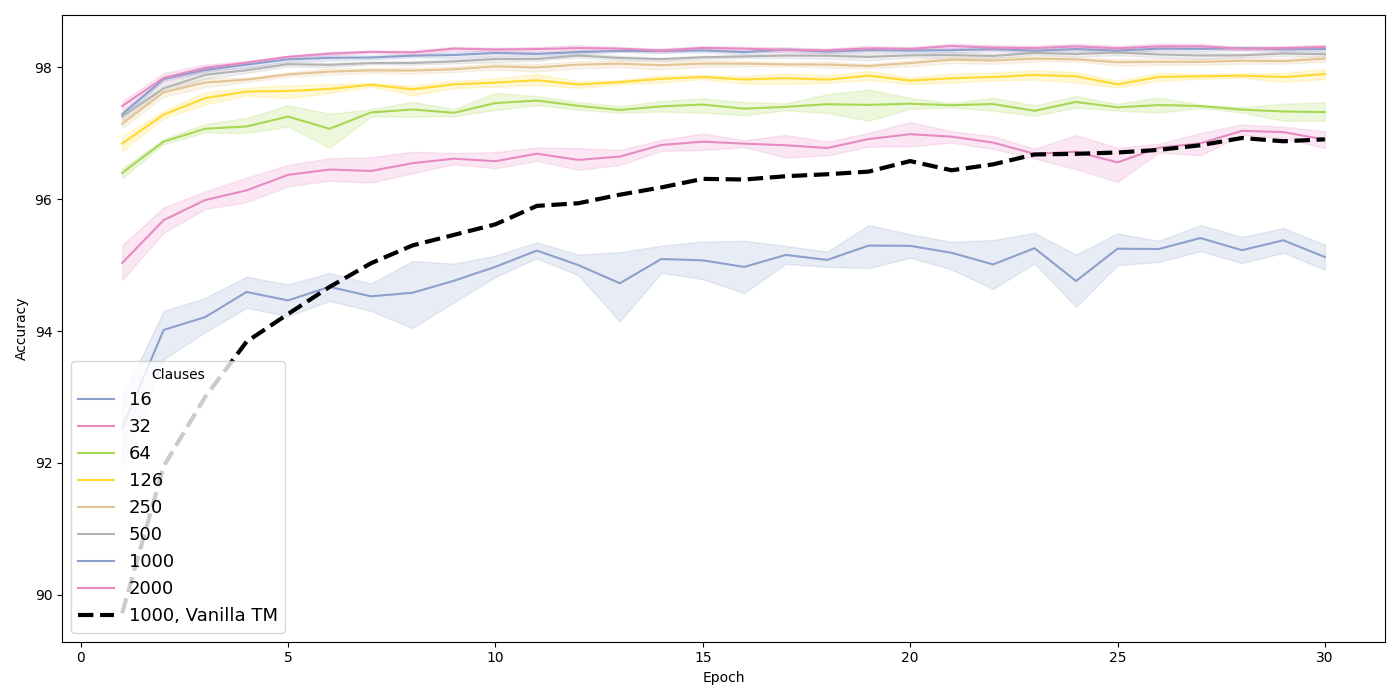}
\caption{Detailed view of performance degradation due to reduced clause count in the HVTM on the MNIST dataset, using \ac{RbE}.}\label{fig:IMG_Clauses_det}
\end{figure}



\section{Conclusion}
\label{sec:conclusion}
In conclusion, this paper introduces the \ac{HVTM}, a novel extension of the \ac{TM} that leverages high-dimensional hypervectors to enhance Booleanization of complex data types and consequently the classification process. The \ac{HVTM} demonstrates an improved capacity for both accuracy and learning speed, with reduced in clause numbers. This has been substantiated via experiments across the domains of text classification, image classification and cheminformatics.

\acp{HVTM}' success in these domains underscores the potential of hyperdimensional computing, in conjunction with \ac{TM}, to effectively manage and interpret complex data structures. 

\textbf{Future Work.}  This paper lays the foundation for further exploration in optimizing hyperspace utilization with Boolean Algebra. Key areas for future work include: 

\textit{Refining the Binding and Bundling of \acp{HV}}: 
These show great potential for encoding features into compact and easy to compute inputs. Not only do they expand the input space by extra dimensions while retaining feature information throughout the learning process, but also allow for constant transparency and interpretability. Leveraging \ac{HV} overlaps can be explored via different combination and improvements on these techniques.

\textit{Expanding HV Applications in Cheminformatics}: 
Currently, morgan fingerprint sub-graph bit positions have been simply converted into \ac{HV} form. \ac{HV} representations of compounds can be expanded upon in greater resolution by including bindings for each sub-substructure graph ECFP encoding and their number. This would allow variable length representations of compounds to be imprinted on fixed length \acp{HV} suitable for the \ac{HVTM}. Similarly, 2D chemical graphs can be converted into \acp{HV} where each unique node (atom) and vertices (bond) is provided a sparse \ac{HV} representation and bound together into sub-graph components. Through leveraging \acp{HV}, variable length graph-based TM models become feasible.

\textit{Exploring Reasoning by Elimination}: 
In current experiments we observe that it creates clauses containing negative hyperliterals which point out what doesn't belong to a certain class rather than focusing on specific hyperliterals which are true for such class. Further research is necessary to explore in detail how it affects the learning and subsequent interpretability across domains.




\nocite{*}
\bibliographystyle{IEEEtran}
\bibliography{References}

\end{document}